\documentclass{article}
\usepackage{spconf}
\usepackage{amsmath}
\usepackage{graphicx}
\usepackage[table,xcdraw]{xcolor}
\usepackage{multirow}
\usepackage{amssymb}
\usepackage{soul}
\usepackage{color}
\usepackage{makecell}
\usepackage{booktabs}
\usepackage{adjustbox}
\usepackage[justification=justified,singlelinecheck=false]{caption}
\usepackage{tabularx}
\newcolumntype{C}{>{\centering\arraybackslash}X}

\title{LibriSpeech-PC: Benchmark for EVALUATION OF PUNCTUATION AND CAPITALIZATION CAPABILITIES OF END-TO-END ASR MODELS}
\name{
    Aleksandr Meister,
    Matvei Novikov,
    Nikolay Karpov,
    Evelina Bakhturina,
    Vitaly Lavrukhin,
    Boris Ginsburg
}

\address{
    NVIDIA, \\ 
    Santa Clara, USA
    }

\copyrightnotice{979-8-3503-0689-7/23/\$31.00~\copyright2023 IEEE}

\begin{document}
%
\maketitle
\begin{abstract}
Traditional automatic speech recognition (ASR) models output lower-cased words without punctuation marks, which reduces readability and necessitates a subsequent text processing model to convert ASR transcripts into a proper format. Simultaneously, the development of end-to-end ASR models capable of predicting punctuation and capitalization presents several challenges, primarily due to limited data availability and shortcomings in the existing evaluation methods, such as inadequate assessment of punctuation prediction. In this paper, we introduce a LibriSpeech-PC benchmark designed to assess the punctuation and capitalization prediction capabilities of end-to-end ASR models. The benchmark includes a LibriSpeech-PC dataset with restored punctuation and capitalization, a novel evaluation metric called Punctuation Error Rate (PER) that focuses on punctuation marks, and initial baseline models. All code, data, and models are publicly available.

\end{abstract}
\begin{keywords}
automatic speech recognition, word error rate, punctuation, capitalization, end-to-end models
\end{keywords}
\section{Introduction}
\label{sec:intro}

Traditional Automatic Speech Recognition (ASR) models typically generate text in lowercase without punctuation. However, practical applications often require properly formatted text to enhance the user experience and optimize performance in downstream models such as Named Entity Recognition (NER) and Neural Machine Translation (NMT) \cite{hrinchuk2022nvidia}. To address this need, a separate punctuation and capitalization model has been commonly employed in a cascaded fashion to post-process ASR output into more readable text \cite{puaics2021capitalization}.
Recent studies have demonstrated \cite{radford2022robust} that end-to-end (E2E) ASR models can effectively handle punctuation and endpoint predictions without the need for an additional punctuation and capitalization (PC) model. A growing trend toward training end-to-end ASR models to predict punctuation and capitalization presents various challenges, such as:
\begin{itemize}
\item Lack of datasets suitable for training E2E ASR models with PC. Conventional ASR models rely on non-capitalized and unpunctuated reference texts; as a result, most publicly-released datasets\footnote{https://openslr.org} contain already processed data and are not applicable for E2E ASR PC model training.
\item Deficiency of evaluation metrics. Word Error Rate (WER) and Character Error Rate (CER) are the primary metrics for evaluating ASR models. Based on edit distance measurement, these metrics show the difference between reference and predicted strings, accumulating errors for any token mismatch. These metrics, along with conventional ones like F1 score, treat all ASR mistakes equally and fail to distinguish errors in words, punctuation marks, and capitalized words separately.
\end{itemize}
This paper presents the LibriSpeech-PC benchmark for evaluating joint ASR models with punctuation and capitalization, incorporating the following components:
\begin{itemize}
\item Punctuation Error Rate: A novel metric for evaluating E2E ASR models with punctuation and capitalization, enabling detailed analysis of punctuation and capitalization errors respectively.
\item LibriSpeech-PC: A dataset based on LibriSpeech \cite{panayotov2015librispeech} with restored punctuation and capitalization.
\item Initial baselines for Cascade and E2E ASR PC English models measured on the proposed benchmark.
\end{itemize}
All code and the LibriSpeech-PC dataset are publicly available to facilitate further advancements in ASR research.

\section{Related Work}
\label{sec:related}

\subsection{Punctuation and Capitalization Prediction}
Papers \cite{shriberg2001can, christensen2001punctuation} were among the first works to combine linguistic and prosodic features for punctuation and capitalization restoration. These features were applied in classical algorithms like decision trees and multi-layer perceptron. The study by \cite{tilk2015lstm} extended this approach by utilizing both audio and text features in an LSTM neural network.

The cascade sequence-to-sequence approach, which is particularly effective in recovering punctuation and capitalization for long ASR text outputs, is detailed in \cite{zelasko2018punctuation}. The transformer-based approach, which translates raw transcriptions to punctuated and capitalized text, is presented in \cite{nguyen2019fast}. These approaches are functional for streaming but necessitate additional maintenance and increase the latency of the pipeline.
Subsequently, \cite{sunkara2020multimodal} demonstrated that joint acoustic and text features pre-trained in a model outperform a text-only model in punctuation and capitalization prediction. However, this still relied on a separate model, requiring additional maintenance and increasing latency.

The work by \cite{guan2020end} introduced an end-to-end acoustic model for punctuation and capitalization tasks. Comparative analyses between the end-to-end and cascade approaches were later conducted \cite{nozaki2022end}, including the proposal of a separate decoder for punctuation \cite{zhou2022punctuation}.

The Whisper paper \cite{radford2022robust} demonstrated that ASR models could be trained to output text with punctuation and capitalization. However, a major drawback of this model is its inability to operate in streaming mode, which is crucial for real-time ASR applications.

\subsection{Metrics for Punctuation and Capitalization}
We consider several known approaches from the literature to evaluate the quality of the predicted text with punctuation and capitalization.

The first approach, which we refer to as F1-based, uses precision, recall, and F1 scores for evaluating text-based models \cite{tilk2015lstm}. These metrics are prevalent in literature for punctuation and capitalization \cite{puaics2021capitalization}. However, in ASR applications, it is often impossible to calculate these scores for all samples due to the absence of certain tokens in either the predicted or ground truth texts, which leads to inaccurate evaluation especially when recognition errors are present.

The second approach involves calculating WER-based metrics, including WER PC (word error rate after both punctuation and capitalization have been predicted), WER C (word error rate after capitalization has been predicted), and the traditional WER  \cite{radford2022robust}. Punctuation errors occur when the model incorrectly predicts the use or placement of punctuation marks. This way, cascade, and end-to-end models can be compared properly, because every sample is utilized.

The third notable approach is distance-based, such as Damerau-Levenshtein Slot Error Rate and Punctuation Specific Damerau-Levenshtein Slot Error Rates \cite{guan2020end}. These metrics can be calculated using the entire dataset.

While there have been significant advancements in punctuation and capitalization prediction, the choice and effectiveness of evaluation metrics remain crucial for meaningful comparisons and benchmarking. This paper aims to address these challenges by introducing a new benchmark and evaluation metric.

\section{Proposed metric}
\begin{figure}[t!]
    \centering
    \includegraphics[width=0.98\columnwidth]{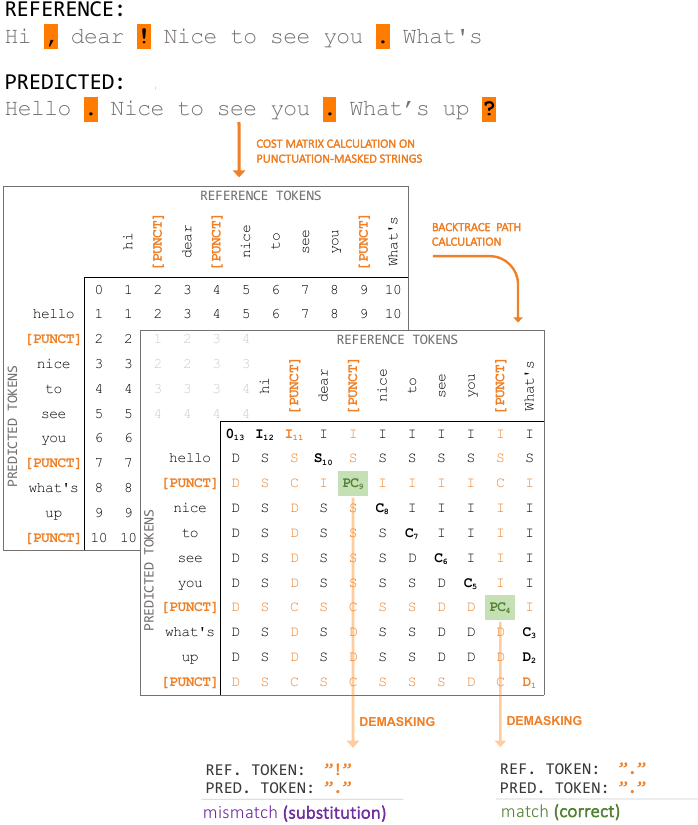}
    \caption{Algorithmic part of the Punctuation Error Rate (PER) calculation: 1. Mask punctuation tokens with a unified label \textit{[PUNCT]}. 2. Construct a cost matrix 3. Backtrace to find the shortest path. 4. Once correct punctuation positions are found, check the correctness of the predicted punctuation tokens, i.e.,  "\textit{Correct}" vs "\textit{Substitution".}}
\end{figure}

\label{sec:proposed}
We introduce the Punctuation Error Rate (PER) metric, which is relevant for evaluating the capabilities of both joint end-to-end ASR models and cascades. This metric facilitates a comparative analysis of these two methodologies in terms of their ability to predict punctuation marks.

The proposed metric is based on the Levenshtein distance algorithm, which uses elementary operations to measure the difference between reference and predicted strings. These elementary operations form the basis for the WER:

\begin{align}
WER = \frac{S + D + I}{N_{ref}} = \frac{S + D + I}{S + D + C}
\end{align}

where $S$ - the number of substitutions, $D$ - the number of deletions, $I$ - the number of insertions, $C$ - the number of matches, $N$ - the number of tokens in the reference sequence.

While the WER aggregates the operations among all tokens of the compared sequences, our metric focuses only on a predetermined subset of tokens, specifically punctuation marks, in the quest to examine punctuation prediction capabilities:

\begin{align}
PER = \frac{I_{P} + D_{P} + S_{P}}{I_{P} + D_{P} + S_{P} + C_{P}}
\end{align}

where each operation is related to the group of punctuation tokens only.
\begin{itemize}
\item $D_{P}$ - instances where a token from the target group is missing at the corresponding position relative to the reference sequence.
\item $I_{P}$ - instances where a token from the target group is superfluous relative to the reference sequence.
\item $S_{P}$ - instances where tokens from the target group exist at identical positions in both the reference and hypothesis sequences but fail to match.
\end{itemize}

The main feature of backtrace matrix calculation is in target tokens masking before computing. In contrast to WER, we use the matrix to estimate the correctness of the target token position only. The masking approach helps to define substitution operation inside the target group instead of naive algorithm substitution operation.

After the position prediction, we perform the check for the exact match of the masked token values.

After calculating the number of correct and substituted tokens we can calculate the number of deletions and insertions. 

\begin{align}
D_{P} = N_{P_{ref}} - (S_{P} + C_{P})
\end{align}
\begin{align}
I_{P} = N_{P_{hyp}} - (S_{P} + C_{P})
\end{align}

where $N_{P_{ref}}$ - amount of punctuation tokens in reference string, $N_{P_{hyp}}$ - amount of punctuation tokens in hypothesis string.

This Punctuation Error Rate equation aligns with the conventional accuracy calculation:

\begin{align}
Accuracy = \frac{TP  + TN}{TP + TN + FP + FN}
\end{align}

The Punctuation Error Rate (PER) and the Word Error Rate (WER) could be used to measure the performance of ASR models both, but they differ in the following aspects: 
\begin{itemize}
\item \textbf{Target Evaluation Element:} While both PER and WER use the Levenshtein distance algorithm to measure the difference between the reference and the hypothesis sequences, they differ in the elements they evaluate. WER focuses on the entire word sequence, including all words and tokens. In contrast, PER concentrates only on a predefined subset of tokens, in this case, punctuation marks.
\item \textbf{Normalization:} In WER calculation, the number of total errors (substitutions, deletions, and insertions) is normalized by the total number of tokens in the reference sequence. In PER, the total number of errors is normalized by the sum of errors and correct predictions. As a result, PER takes values between 0 and 1 and can be interpreted similarly to a traditional accuracy measure.
\item \textbf{Application:} WER is a general-purpose metric used for evaluating the overall performance of ASR systems. PER, on the other hand, is particularly useful for a more granular evaluation of a model's performance, especially its ability to predict a specific subset of tokens, such as punctuation marks.
\end{itemize}

\section{LibriSpeech-PC Dataset}
\label{sec:LibriSpeech-PC}

We curated a version of the LibriSpeech dataset\cite{panayotov2015librispeech}, named LibriSpeech-PC, to include punctuation and capitalization information for training and evaluating ASR with PC models. To recover punctuation and capitalization in the LibriSpeech dataset, we 1) traced the original text sources of the LibriSpeech audiobooks and then 2) aligned the original LibriSpeech transcripts with the text from the book sources. Both sample segmentation and subset division from the original LibriSpeech dataset were preserved. 

During the alignment process, we dropped samples when one of the following conditions was present:
\begin{itemize}
    \item Samples contained non-standard Unicode characters, resulting in a mismatch between the original text and LibriSpeech transcripts. For example, the name "Kutúzov" in the original text is not equal to "Kutuzov" in LibriSpeech, this would cause an alignment issue.
    \item If the recovered text and the original text deviated by more than two words, it signaled possible restoration errors. This was particularly common with specific phrases, such as "CHAPTER II".
    \item If the sample duration was less than a second, the sample typically only contained a single word and could create confusion. This was due to the inconsistent application of capitalization of the first letter and punctuation at the end of these short samples. For instance, one could encounter a sample like "One" or "one.".
    \item If a sample was entirely in uppercase, which is typically due to an error in the original book's text. An example of this could be a sentence like "JULIE WAS GOING TO THE WRONG HOUSE."
\end{itemize}

The total duration for each split in hours is presented in Table \ref{tab:dataset_splits} with a retention rate above 90\%. 
\begin{table}[t]
\caption{During the PC restoration process, we dropped some samples. There is a comparison of the duration of LibriSpeech and LibriSpeech-PC datasets.}
\label{tab:dataset_splits}
\resizebox{\columnwidth}{!}{
\begin{tabular}{lccc}
        \toprule
        \textbf{Subset} & \makecell{\textbf{Original LS} \\ \textbf{Duration (hrs)}} & \makecell{\textbf{LS PC} \\ \textbf{Duration (hrs)}} & \makecell{\textbf{Retention} \\ \textbf{Rate (\%)}} \\
        \midrule
        train-other-500 & 496.86 & 449.10 & 90.38 \\ 
        train-clean-360 & 363.31 & 333.13 & 91.69 \\
        train-clean-100 & 100.59 & 91.61 & 91.07 \\
        dev-clean & 5.39 & 4.96 & 92.02 \\
        dev-other & 5.12 & 4.77 & 93.16 \\
        test-clean & 5.40 & 4.98 & 92.22 \\
        test-other & 5.34 & 5.17 & 96.81 \\
        \bottomrule
\end{tabular}
}
\end{table}

\begin{table*}[t]
\centering
 \caption{Number of samples where WER equals to zero and their corresponding quality metrics (F1, Precision, Recall) for capital letters (Capitalization) and punctuation marks (Punctuation)}
 \label{tab:results_number_of_samples_for_f1}
    \begin{tabular}{lc|ccc|ccc|ccc}
    \toprule
      &  & Total & WER=0 &  Inter- & \multicolumn{3}{c|}{Capitalization} & \multicolumn{3}{c}{Punctuation} \\
     \textbf{Setup} & \textbf{Subset} & samples & samples & section & F1 & Prec. & Recall & F1 & Prec. & Recall \\

     \midrule
     Tuned Cascade & test-clean & 2417 & 1618 (66.94\%) & 375 & 94.47 & 95.19 & 93.78 & 84.75 & 83.95 & 85.60 \\
     E2E Conformer & test-clean & 2417 & 1605 (66.40\%) & (15.5\%) & 94.85 & 95.53 & 94.20 & 87.40 & 86.11 & 88.90 \\
     \midrule
     Tuned Cascade & test-other & 2856  & 1376 (48.17\%) & 1119 & 94.53 & 95.87 & 93.29 & 83.40 & 84.41 & 82.70 \\
     E2E Conformer & test-other & 2856 & 1439 (50.38\%) & (39.1\%) & 95.50 & 96.37 & 94.67 & 88.27 & 86.87 & 89.93 \\
     \bottomrule
    \end{tabular}
\end{table*}

\section{Experiments}
\label{sec:experiments}

LibriSpeech-PC contains a wide variety of punctuation marks and word casing forms, and for ASR model training, we perform the following prepossessing of the transcripts:
\begin{enumerate}
  \item Only periods, commas, and question marks are retained, while other punctuation marks have been removed due to their limited presence in the dataset.
  \item Words and punctuation marks were separated with a space. For instance, ``done.” was transformed to ``done . ”.
  \item Multiple consecutive whitespaces were replaces with a single whitespace.
\end{enumerate}


We designed a set of experiments to compare end-to-end models with the cascade approach where an automatic speech recognition model outputs lower cased words without punctuation marks followed by a text-to-text punctuation model.

\subsection{Base and Tuned Cascades}
\label{subsec:CascadeModelConfig}

Our \textbf{Base Cascade} approach consists of a Conformer-based \cite{gulati2020conformer} ASR model \texttt{stt\_en\_conformer\_ctc\_large\_ls}\footnote{https://catalog.ngc.nvidia.com} followed by a BERT-based \cite{devlin2018bert} punctuation and capitalization model \texttt{punctuation\_en\_bert} with dual heads where each head predicts punctuation and capitalization per token.

In \textbf{Tuned Cascade} variant, we finetune pretrained BERT-based PC model from the Base Cascade for 3 epochs with a batch size of 15000 tokens per GPU. We use Adam optimizer \cite{kingma2014adam} with a learning rate of $1 \times 10^{-4}$ and a warmup annealing scheduler set to 10\% of the total steps.

\subsection{E2E ASR with Punctuation and Capitalization}
\label{subsec:EndToEndAcousticModelConfig}

We consider two end-to-end ASR models which able to predict text with PC: \textbf{Whisper-base} model\footnote{https://huggingface.co/openai/whisper-base} and our \textbf{Conformer E2E} model based on a Conformer acoustic model. Conformer E2E model was fine-tuned with punctuation and capitalization from the publicly available checkpoint \texttt{stt\_en\_conformer\_ctc\_large\_ls} over a course of 100,000 steps, with a global batch size of 2048 samples. We use AdamW optimizer \cite{loshchilov2017decoupled} with a learning rate of $0.001$, $\beta_1 = 0.9$ and $\beta_2 = 0.98$, a weight decay of $0.001$, and a cosine annealing scheduler with 10,000 warm-up steps and a minimum learning rate of $1 \times 10^{-6}$. The SentencePiece Unigram tokenizer \cite{kudo2018subword} was used with a vocabulary of 128 tokens.

\addtolength{\tabcolsep}{-1pt} 
\begin{table*}[t]
 \caption{Experiment results for baseline and our \textbf{Cascade} and \textbf{E2E} approaches on LibriSpeech-PC test-clean and test-other sets using WER-based metrics}
 \label{tab:results_word_error_rates_clean}
    \begin{tabularx}{0.99\textwidth}{ll|rrrr|rrrr}
    \toprule
     & Num. & \multicolumn{4}{c|}{LibriSpeech-PC test-clean (\%)}  & \multicolumn{4}{c}{LibriSpeech-PC test-other (\%)}  \\
     \textbf{Setup} & param. & \textbf{PER} & \textbf{WER} & \textbf{WER C} & \textbf{WER PC}  & \textbf{PER} & \textbf{WER} & \textbf{WER C} & \textbf{WER PC} \\
     \midrule
     Whisper-base & 74M  & 36.97  & 4.50  & 6.82  & 11.23   & 40.27  & 11.23  & 13.57  & 17.82  \\
     Base Cascade & 115+109M & 38.38  & 2.60  & 4.83  & 26.36  & 38.60 & 5.79 & 7.99 & 29.96 \\
     Tuned Cascade & 115+109M & 36.60  & 2.86   & 5.09   & 9.45 & 37.42  & 6.26  & 8.66  & 12.90 \\
     E2E Conformer & 115M & 29.48  & 2.83   & 4.72   & 8.11  & \textbf{27.37}  & 5.65  & 7.57  & 10.48 \\
     
     \midrule
     E2E Conformer LS LM & 115M & 29.27  & 2.22  & 4.13  & 7.66  & 28.44  & 4.69 & 6.65  & 9.80 \\
     E2E Conformer CC LM & 115M & 32.40  & 2.24  & \textbf{4.04}  & 8.00  & 31.78  & 4.71 & \textbf{6.54}  & 10.36  \\
     E2E Conformer LS+CC LM & 115M & \textbf{29.24}  & \textbf{2.22}  & 4.05  & \textbf{7.66} & 27.67  & \textbf{4.67} & 6.55 & \textbf{9.80} \\
     \bottomrule
    \end{tabularx}
\end{table*}
\addtolength{\tabcolsep}{1pt}

\subsection{E2E ASR with n-gram Language Models}
\label{subsec:LanguageModel}
The end-to-end automatic speech recognition model can be improved using beam-search algorithm with a language model (LM) \cite{karpov21_interspeech}. We build a word-piece level N-gram language model with byte-pair-encoding (BPE) tokens using SentencePiece\footnote{https://github.com/google/sentencepiece} and KenLM\footnote{https://github.com/kpu/kenlm} toolkits from transcription of LibriSpeech-PC. Token sets of our ASR and LM model match.

To enrich our LM model with more diverse examples of punctuation and capitalization, we download a random subset from the Common Crawl dataset \cite{luccioni-viviano-2021-whats}. The Common Crawl subset is 100 times bigger than LibriSpeech-PC and is normalized and cleaned similarly to LibriSpeech-PC. 

We experiment with three 7-gram language models in ARPA format:
\begin{itemize}
    \item Transcriptions of LibriSpeech-PC dataset (\textbf{LS});
    \item Random subset of the Common Crawl dataset (\textbf{CC});
    \item Mix of LibriSpeech-PC and Common Crawl subset with equal weights (\textbf{LS+CC}).
\end{itemize}
To combine two N-gram models with equal weights we used OpenGrm NGram\footnote{https://www.opengrm.org/twiki/bin/view/GRM/NGramLibrary} library. As expected, E2E model with beam-search decoder and language model performs better than the pure one, see Section \ref{results}.

\section{Results}
\label{results}

\subsection{F1-based Metrics}
\label{subsec:TraditionalMetrics}
As discussed in Section \ref{sec:related}, the F1 score could be easily used for evaluation only on samples where WER is zero. To follow  previous works we compare Tuned Cascade and our E2E Conformer model using F1, Precision and Recall using capital letters (Capitalization) and punctuation marks (Punctuation). The rest models were excluded from comparison to reduce the number of measurements. Only 48-66\% of samples from test-clean and test-other have WER=0 in our two experimental setups. As you can see in Table \ref{tab:results_number_of_samples_for_f1} the E2E model always outperform Tuned Cascade.

For a fair comparison of the models we should consider only the intersection of those zero-WER sample subsets, which could potentially reduce the evaluation set even further. Test-clean set has 375 (15.5\%) of intersected samples, test-other has 1119 (39.1\%). These intersection sizes could vary depending on the quality of the model, which makes the metrics not reproducible.


\subsection{WER-based Metrics}
\label{subsec:new_metrics}

Another approach to evaluate and compare the PC model is word error rate variations. 
The Table \ref{tab:results_word_error_rates_clean} shows the results of our speech recognition setups on the LibriSpeech-PC test-clean and test-other datasets, respectively. The models are evaluated based on their ability to transcribe the audio into text with PC. 
There are a number of metrics: word error rate with no punctuation and no capitalization (\textbf{WER}), word error rate with capitalization and no punctuation (\textbf{WER C}), word error rate with punctuation and capitalization (\textbf{WER PC}) and punctuation errors (\textbf{PER}) made by the model.

The Table \ref{tab:results_word_error_rates_clean} summarize the performance of the cascade and end-to-end models with and without language model rescoring. On the test-clean subset the E2E model with the joint language model (LibriSpeech-PC and Common Crawl) achieves the best overall performance with a WER PC of 7.66\% and PER of 29.24\%. The E2E model without language model has a slightly higher WER PC of 8.11 \% and PER of 29.48\%, while the Tuned Cascase of models has a higher PER of 36.6\% and a WER PC of 9.45\%. The Base Cascade models has the highest WER PC of 26.36\%, followed by Whisper-base model with results 11.23. The similar situation can be fount with the test-other subset.

\begin{figure}[t!]

    \centering
    \includegraphics[width=\columnwidth]{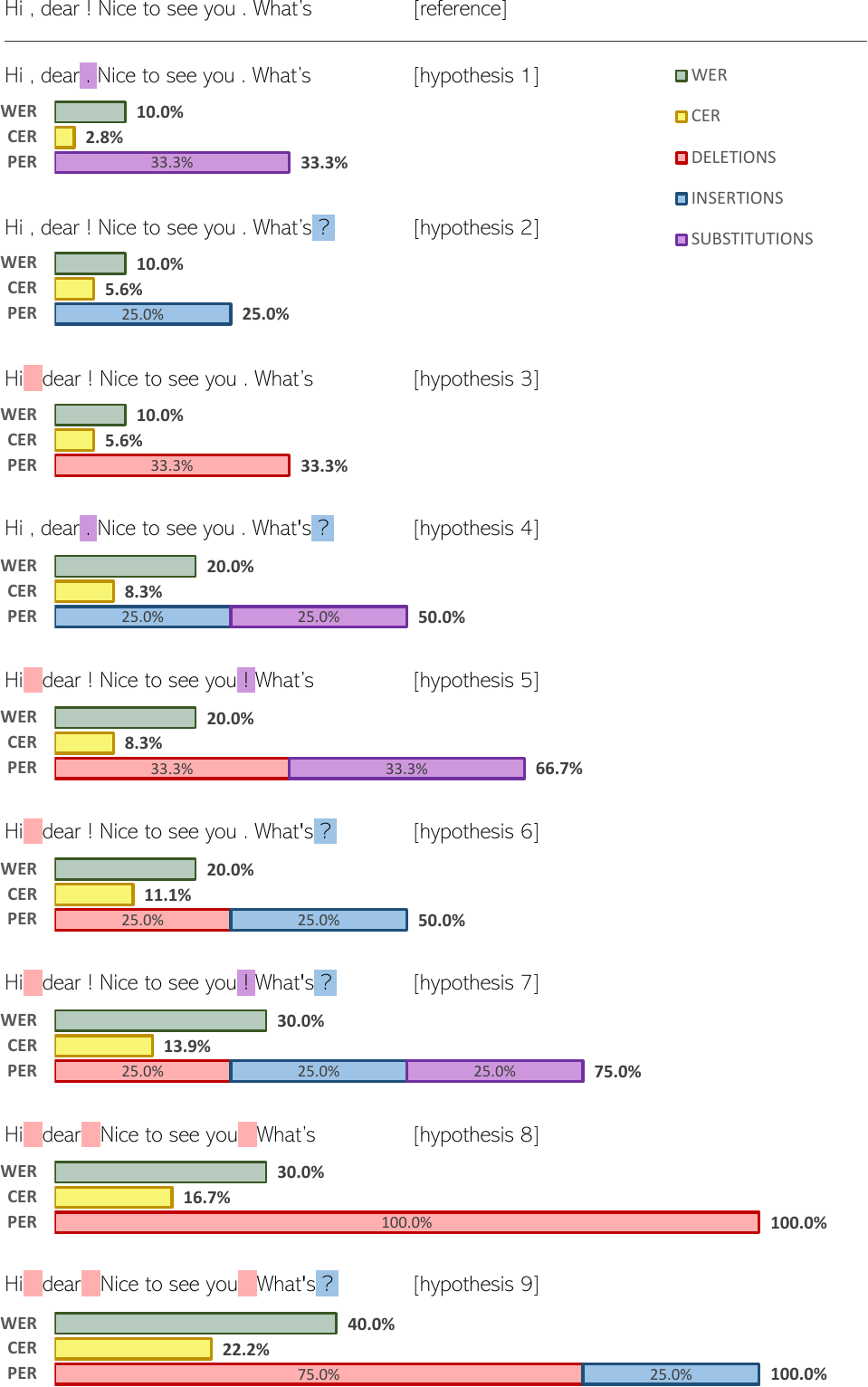}
    \caption{Sensitivity of Word Error Rate (WER), Character Error Rate (CER), and Punctuation Error Rate  (PER) to various punctuation errors.}
    \label{fig:per}
\end{figure}

Let's consider some examples to demonstrate why the PER metric is needed and how it can be beneficial in evaluating punctuation prediction:

\textbf{Example 1:} For the reference text ''I was done .'' and the hypothesis ''I was done'', the PER is 100.0\% whereas the WER PC is only 25.0\%. The high PER effectively captures the complete absence of punctuation, which may not be as apparent when solely relying on WER PC.

\textbf{Example 2:} For the reference text ``Let's eat , Bob !'' and the hypothesis ``Let's eat Bob !'', the PER is 50.0\% while the WER PC is 20.0\%. This example highlights how crucial punctuation can be in conveying the intended meaning. The PER is effective in capturing this critical punctuation error.

These examples highlight the ability of PER to provide a more focused evaluation of punctuation prediction. Traditional WER-based metrics may not sufficiently reflect the significance of punctuation errors, which can be crucial for meaning and clarity. As such, using PER in conjunction with WER offers a more comprehensive and nuanced evaluation of a model's performance in predicting punctuation.

The Fig. \ref{fig:per} demonstrates one more example of error rate metrics comparison between a reference string and hypotheses containing diverse types punctuation-related errors. WER and CER are only minimally effective in evaluating punctuation errors, as they account for errors across all tokens. In contrast, PER only considers errors in punctuation marks, thereby providing an isolated evaluation of their prediction quality.

\addtolength{\tabcolsep}{-4pt} 
\begin{table} 
\centering
 \caption{Error Rates on \textbf{LibriSpeech test-clean} set}
 \label{tab:results_punct_error_rates_clean}
 \begin{tabularx}{\columnwidth}{lCCC}
    \toprule
     \textbf{Metrics} & \textbf{Period} & \textbf{Comma} & \textbf{Question Mark} \\
     \midrule
     Correct & 76.0\% & 66.73\% & 84.02\% \\
     Deletions & 2.52\% & 13.53\% & 3.2\% \\
     Insertions & 10.86\% & 18.23\% & 5.94\% \\
     Substitutions (total) & 10.63\% & 1.5\% & 6.85\% \\
     \midrule
     Substitutions (to period) & - & 1.15\% & 2.74\% \\
     Substitutions (to comma) & 9.93\% & - & 4.11\% \\
     Substitutions (to QM) & 0.7\% & 0.35\% & - \\
     \midrule
     PER & 24.01\% & 33.26\% & 15.99\% \\
     \bottomrule
    \end{tabularx}
\end{table}

\section{Conclusions}
\label{sec:conclusions}

We introduced the Punctuation Error Rate (PER), an innovative metric designed to evaluate punctuation and capitalization recovery in automatic speech recognition (ASR) systems. Unlike established metrics such as F1 and WER, the PER operates independently of words and thus provides a more suitable evaluation for both end-to-end and cascade models. 
Moreover, the granular level of detail provided by PER, particularly with respect to each individual punctuation mark (comma, period, and question mark), makes it an advantageous tool for both researchers and practitioners.
The development and application of the PER in this study underscores the necessity for more nuanced evaluation techniques as language models continue to evolve and become increasingly sophisticated. By focusing specifically on punctuation and capitalization recovery, PER contributes a fresh perspective to ASR model evaluation.

In addition to this, we developed the LibriSpeech-PC dataset, an adaptation of the widely used LibriSpeech corpus, enhanced with punctuation and capitalization. We anticipate that this novel dataset will be instrumental for future research in this specific domain. 
Our empirical findings highlight that the end-to-end Conformer-based ASR model, when trained on the LibriSpeech-PC dataset, shows superior performance compared to the cascade model. This insight could influence future model selection and design in punctuation and capitalization recovery tasks.

In closing, our research presents a compelling argument for the inclusion of the PER as a standard evaluative metric in the field of ASR. We believe that this work will influence further advancements in ASR systems, contributing to an enhanced understanding of performance metrics and facilitating progress in automatic speech recognition.

\bibliographystyle{IEEEbib}
\bibliography{refs}
\end{document}